\documentclass[preprint,12pt]{elsarticle}

\usepackage{graphicx}
\usepackage{amssymb}
\usepackage{amsmath}
\usepackage{url}
\usepackage{diagbox}

\usepackage{lineno}
\usepackage[table]{xcolor}
\usepackage{multirow}

\newcommand{\x}{{\bf x}}

\newcommand{\z}{{\bf z}}
\newcommand{\w}{{\bf w}}

\newcommand{\beq}{\begin{equation}}
\newcommand{\eeq}{\end{equation}}
\newcommand{\bde}{\begin{definition}}
\newcommand{\ede}{\end{definition}}
\newcommand{\bpp}{\begin{property}}
\newcommand{\epp}{\end{property}}
\newcommand{\bpr}{\begin{proposition}}
\newcommand{\epr}{\end{proposition}}
\newcommand{\bex}{\begin{example}}
\newcommand{\eex}{\end{example}}
\newcommand{\bco}{\begin{corollary}}
\newcommand{\eco}{\end{corollary}}
\newcommand{\bre}{\begin{remark}}
\newcommand{\ere}{\end{remark}}
\newcommand{\bal}{\begin{algorithm}}
\newcommand{\eal}{\end{algorithm}}
\newcommand{\ble}{\begin{lemma}}
\newcommand{\ele}{\end{lemma}}

\definecolor{amber(sae/ece)}{rgb}{1.0, 0.49, 0.0}
\definecolor{amberlite(sae/ece)}{rgb}{1.0, 0.69, 0.2}
\newcommand{\hl}[1]{{\color{black}{#1}}}
\newcommand{\hlsec}[1]{{\color{black}{#1}}}

\journal{}

\begin{document}

\begin{frontmatter}


\title{Evaluation of machine learning algorithms for Health and Wellness applications: a tutorial}



\author[uef]{Jussi Tohka}
\author[tuni]{Mark van Gils}
\address[uef]{A.I. Virtanen Institute for Molecular Sciences, University of Eastern Finland, Kuopio, Finland, email: jussi.tohka@uef.fi }
\address[tuni]{Faculty of Medicine and Health Technology, Tampere University, Sein{\"a}joki, Finland, email: mark.vangils@tuni.fi }

\begin{abstract}
Research on decision support applications in healthcare, such as those related to diagnosis, prediction, treatment planning, etc., has seen strongly growing interest in recent years. This development is thanks to the increase in data availability as well as advances in artificial intelligence and machine learning research. Highly promising research examples are published daily. However, at the same time, there are some unrealistic, often overly optimistic, expectations and assumptions with regards to the development, validation and acceptance of such methods. The healthcare application field introduces requirements and potential pitfalls that are not immediately obvious from the 'general data science' viewpoint. Reliable, objective, and generalisable validation and performance assessment of developed data-analysis methods is one particular pain-point. This may lead to unmet schedules and disappointments regarding true performance in real-life with as result poor uptake (or non-uptake) at the end-user side. It is the aim of this tutorial to provide practical guidance on how to assess performance reliably and efficiently and avoid common traps especially when dealing with application for health and wellness settings. Instead of giving a list of do's and don't s, this tutorial tries to build a better understanding behind these issues and presents both the most relevant performance evaluation criteria as well as approaches how to compute them. Along the way, we will indicate common mistakes and provide references discussing various topics more in-depth. 
\end{abstract}

\begin{keyword}
Machine learning \sep Artificial intelligence \sep Biomedicine \sep Life Sciences \sep Performance Assessment \sep Decision Support Systems


\end{keyword}

\end{frontmatter}



\section{Introduction}
\label{sec:introduction}
Data-driven approaches for healthcare decision support, such as those making use of Machine Learning (ML), have seen a surge in interest over recent years, partly driven by the promising results that a 'reborn' artificial intelligence (AI) research branch has generated. As the name says, these approaches rely on the availability of data to extract knowledge and train algorithms. This is opposed to, e.g., modeling approaches in which physiological, physics-based, mathematical, and other equations form the basis of algorithms, or, rule-based systems in which reasoning processes are obtained by translating domain-experts' knowledge into computer-based rules.

Focusing on data-driven systems, the data plays a role in several components during the development and actual usage phases. First, we need data to extract knowledge from, i.e., to develop and train algorithms so that they learn-by-example the properties of the problem at hand and get better at solving the problem by repeatedly providing example data. Second, we need to monitor during the development phase how promising the algorithms are and make choices, e.g., concerning optimisation of parameters or choosing different ML paradigms. Methods that do not perform well at all can be discarded, and ones that seem promising can be further optimised. To assess how promising a specific method is, we need to examine how it performs on data that was not used during training.
Finally, to objectively assess how well the final 'best' system performs, we need to apply completely new data to it that has not been used at all thusfar during the research and development process.

Thus, there are at least three stakeholders that have the interest to get as large part of the data pie as possible. 1) the algorithm developer, to get as good method development as possible; 2) the validation assessment responsible, who would help to steer the development process as good as possible; and 3) the decision-maker who wants to have as accurate as possible assessment of the merits of the system. Thus, we need to make a trade-off and think about efficient usage of the precious data, and need to carefully define what we truly mean with performance evaluation and what kind of measures or yardsticks would be appropriate when.  

It is the aim of this paper to provide practical guidance on how to assess performance reliable and efficiently and avoid common traps. While the abovementioned phases of research and development and different stakeholder roles are applicable to a wide range of applications in which AI/ML methods are used, the health and wellness domain have own requirements with consequences that are not necessarily immediately grasped when applying 'generic' data-analysis methods to the problems in the field. 

\hl{Information sources are plenty, and indeed, topics covered in this article can are partly discussed elsewhere also --- dispersed over websites, book chapters, articles, blog posts, and different courses. However, the fact that bits of information are spread over different places and that the content-levels of these sources are highly variable and assume different forms of background knowledge makes getting acquainted with the field cumbersome and time-consuming for many. Based on experiences gained in over 25 years in the field the authors were motivated to compile practical knowledge about caveats and best-practices into one single place.
Target audiences include healthcare professionals (ranging from senior clinicians to specialising young doctors), {R\&D} experts in industry, but also to students in biosciences, biomedical engineering and other disciplines for whom this type of information is not part of the standard curriculum.} 

\subsection{Decision Support in Healthcare settings}
Decision support in healthcare is not a new field by itself. For over 50 years the possibilities of computer-based systems to assist healthcare professionals in decision making have been investigated, and numerous prototypes and actual products have been deployed in real-life. There are many decisions that a computer-based algorithm might help with. Perhaps the most common example is that of diagnosis of a disease based on, e.g., medical images or signals. This is a classification task (diseased vs. healthy-case, or disease A vs. disease B). In this case, the users are healthcare professionals (e.g., medical doctors or radiologists). Other common decision-making tasks include risk assessment (the risk of developing a disease or adverse events), predicting hospital resource needs, treatment outcome, helping to plan interventions (making surgery or treatment plans), or monitor a patient state over time to see if a treatment has success or not. 
\hl{Literature includes a large body or research papers evolving from modelling and expert systems approaches to statistical pattern recognition to purely data-driven (AI/ML) approaches. Overview works include \cite{Steyerberg_ClinicalPredictionModels} for a thorough treatment of the case of prediction and statistical modelling approaches, \cite{berner_clinical_2007} is more towards clinical settings, and \cite{topol_deep_2019} for a wide overview of more recent trends and outlooks, especially in the context of developments in AI.}

\hl{From a data science point of view there are many steps that contribute to the processing chain that underlies decision support tools. These include methods, at the 'early parts' of the chain, methods that validate input data, detect artefacts, deal with invalid/missing data, improve signal-to-noise ratio, and extract and select features. They are typically followed by higher-level classification methods or regression methods that provide the 'final output' to the user. How to assess reliably and objectively how good this 'final output' is in practice (i.e., the performance of the system), is the question that this paper addresses.} 

We concentrate primarily on assessing the final outputs of supervised classification and regression approaches as they are an essential part of most decision support systems in healthcare in practice. Unsupervised approaches, such as clustering methods, are often more related to early-stage exploratory research, typically involving visualisation. This is an important step to understand the problem at hand, but accurately quantifying 'performance' in terms of numbers is often less needed. 

\subsection{Further Acceptance Criteria for Decision Support in Healthcare}
It should be noted that, next to performance \emph{per se}, there are many other criteria that influence whether an algorithm will find successful uptake in healthcare practice. Criteria include: 
\begin{itemize}
    \item performance measures such as (classification) accuracy; sensitivity, specificity and others;
    \item usability of methods for end-users (who can be healthcare professionals or patients);
    \item ease of integration in existing processes and workflows;
    \item compatibility and ease of integration with existing IT infrastructures and standards;
    \item robustness (e.g., deal with missing and poor quality data - especially relevant in health data);
    \item explainability and understandability of results (for decision support in healthcare especially, black box solutions are not acceptable); and
    \item proof of actual impact, measured by cost-effectiveness, clinical usefulness, measurable productivity increase, or effects on quality-of-life.
\end{itemize}

In this paper, we focus on the \emph{performance measures}, due to their crucial role in, e.g., diagnostics, risk assessment and treatment planning. However, the other criteria are not less important and would deserve separate discussion in dedicated tutorials.

\section{Supervised classification}
\label{sec:supervised_classification}

This section reviews the background on supervised classification from a more formal viewpoint and introduces the notation. For clarity, we will focus on classification problems, but the treatment of regression problems would be very similar.  For this tutorial, this section sets the scene by introducing the Bayes classifier, exemplifies how the Bayes classifier is approximated, illustrates how the Bayes classification rule depends on the prior probability of classes, and informs about sampling issues when training discriminative classifiers. 

\subsection{General framework and notation}

A classification task is to assign an object (in health applications, usually a person) described by a feature vector $\x = [x_1,\ldots,x_d]$ (e.g., measured blood pressure levels, total cholesterol, age, sex) 
to one of the $c$ classes (e.g., cardiovascular disease in the future or not). The $c$ classes are denoted simply as $1,\ldots,c$. A classifier is represented by a function $\alpha$ that takes as an input a feature vector $\x$ and outputs the class of the object represented by that feature vector. In more practical terms, a classifier outputs $c$ values representing each class, and the class of the highest (or lowest) value is the one selected.  In supervised learning, this function $\alpha$ is constructed based on training data, which is given by pairs $(\x_i,y_i)$, where $\x_i$ is the feature vector of an object belonging to the class $y_i \in \{1,\ldots,c\}$. We assume to have $N$ such pairs, constituting the training data $\{(\x_i,y_i)|i = 1,\ldots,N\}$.

\subsection{Bayes classifier}

\hl{\emph{The Bayes classifier} is the optimal classifier that can be only constructed when all the statistical characteristics of a given classification problem are known.  It is a theoretical construct that is useful in theoretical studies in supervised learning, for deriving practical learning algorithms with some optimality guarantees,  and to understand the concept of the generalization error whose estimation this tutorial is concerned with. In practice, the complete and accurate the statistical characterisation of the classification problem is not known and the classifier must be learned based from training data. However, nearly all the practical classifiers approximate the Bayes classifier in some sense. }

To build the Bayes classifier, we assume to  know the 
\begin{enumerate}
\item {\em prior probabilities} 
$P(1), \ldots, P(c)$ of the classes and 
\item the {\em class
  conditional probability density functions  (pdfs)}
$p(\x|1), \ldots, p(\x|c)$. 
\end{enumerate}
The prior probability $P(j)$
defines what percentage of all objects belong to the class 
$j$. The class conditional pdf $p(\x|j)$
defines the pdf of the feature vectors belonging to $j$. 
Obviously,
$ \sum_{j = 1}^c P(j) = 1$. 

The Bayes classifier is defined as
\begin{equation}
\alpha_{Bayes}(\x) = \arg \max_{j = 1, \ldots,c}
P(j|\x),
\end{equation}
where $P(j|\x)$ is the posterior probability of the class $j$ being the correct class for the object with the feature vector $\x$. 
\footnote{The notation $\arg \max_{\x} f(\x)$ means the value of the argument $\x$
that yields the maximum value for the function $f$. For example, if $f(x) = -(x
- 1)^2$, then $\arg \max_x f(x) = 1$ (and $\max f(x) = 0$).}  In other words,
the Bayes classifier selects the most probable class when the observed
feature vector is $\x$. 

{\em The posterior probability} $P(j|\x)$ is evaluated 
based on the Bayes rule, i.e., $P(j|\x) = \frac{p(\x|j) P(j)}{p(\x)}$. However, $p(\x)$ is equal for all classes and it can be dropped. The Bayes classifier can now be rewritten as   \begin{equation}
  \label{eq:bayes2}
\alpha_{Bayes}(\x) = \arg \max_{j = 1, \ldots,c}
p(\x|j) P(j),
\end{equation}
i.e., the Bayes classifier computes the product of the class conditional density at $\x$ and the prior for class $j$.  

By its definition the Bayes classifier minimizes the conditional
error  
$E(\alpha(\x)|\x) = 1 - P(\alpha(\x)|\x)$ for all $\x$. 
Because of this and basic properties of integrals, the Bayes
classifier also minimizes the classification error 
\begin{equation}
E(\alpha) = \int_\x E(\alpha(\x)|\x) p(\x) d\x. 
\end{equation}
In the above equation, it is important to note that the integration is over {\em all} possible feature vectors, not just those in the training set.  This is the generalization error that we are interested in estimating in the coming sections.   

The classification error $E(\alpha_{Bayes})$ of the Bayes classifier is
called {\em the Bayes error}. It is the smallest possible
classification error for a fixed classification problem.  As mentioned previously, the Bayes classifier is a theoretical construct: in practice we never know the class conditional densities $p(\x|j)$ or class priors $P(j)$ 

We remark that the definition of the Bayes classifier does not
require the assumption that the class conditional pdfs are Gaussian
distributed. The class conditional pdfs can be any proper pdfs. 

\subsection{Bayes formula example (Cancer test)}
It is important to understand the role of the prior probabilities when designing classification rules or considering the strength of evidence.  This equals to understanding of the Bayes formula. We give here a brief example from a healthcare setting, which is summarized from  \footnote{\url{http://yudkowsky.net/rational/bayes}} and \footnote{\url{https://betterexplained.com/articles/an-intuitive-and-short-explanation-of-bayes-theorem/}}, but similar examples appear in various text books on statistics. 

The example is based on the following scenario: 

\begin{itemize} 
\item 1\% of women at age forty who participate in routine screening have breast cancer, i.e., $P(Cancer+) = 0.01$, and therefore 99\% do not, i.e., $P(Cancer-) = 0.99$; 
\item 80\% of mammograms detect breast cancer when it is there (and therefore 20\% miss it), i.e. $P(Test+|Cancer+) = 0.80$, $P(Test-|Cancer+) = 0.20$;
\item 
9.6\% of mammograms falsely detect breast cancer when it is not there (and therefore 90.4\% correctly return a negative result), i.e., $P(Test+|Cancer-) = 0.096$,$P(Test-|Cancer-) = 0.904$ . 
\end{itemize} 
A woman in this age group had a positive mammography in a routine screening.  What is the probability that she actually has breast cancer? 

The correct answer is 7.8\%, an answer that can be quite counter-intuitive at first sight. The answer is obtained based on the Bayes rule. First, note that the question asks for the posterior probability of breast cancer given that the test was positive $P(Cancer+|Test+)$. This probability was not provided above and it must be computed based on the Bayes rule:
$$
P(Cancer+|Test+) = \frac{P(Test+|Cancer+)P(Cancer+)}{P(Test+)} = \frac{0.80 \cdot 0.01}{0.10304} = 7.8\%.
$$
Note that $$
P(Test+) = P(Test+|Cancer+)P(Cancer+) + P(Test+|Cancer-)P(Cancer-). $$

\subsection{Sampling issues}

The training data may be sampled in two distinct ways and it is important to make a distinction between these. In {\em
  mixture (or random) sampling}, the training data is collected for all classes simultaneously conserving the class-proportions occurring in real-life.  In {\em separate sampling}, the training data for
each class is collected separately. For the classifier training, the
difference of these two sampling techniques is that for the mixed
sampling we can estimate the priors $P(1), \ldots, P(c)$ as
\begin{equation}
\hat{P}(j) = \frac{n_j}{\sum_{k = 1}^c n_k}, 
\end{equation}
where $n_j$ is the number of samples from the class $j$. 
On the other hand, the prior probabilities cannot be deduced based on
the separate sampling. Most sampling in healthcare-related studies is separate sampling, whereas most standard textbooks assume mixture sampling. A good overview of the challenges caused by the separate sampling is \cite{shahrokh2013}.    

\subsection{Practical classifiers}

This subsection briefly explains how to construct classifiers based on the training data $\{(\x_i,y_i)|i = 1,\ldots,N\}$. We can approach this problem in several ways. The conceptually simplest way is to approximate the Bayes classifier via generative plug-in classifiers. These approximate the prior probabilities $P(j)$ and the class conditional densities $p(\x|j)$ by estimates $\hat{P}(j)$, $\hat{p}(\x|j)$ found based on training data and substitute \hl{(or plug-in)} these estimates to the formula of the Bayes classifier. The estimates for class conditional pdfs can be either parametric (e.g., Naive Gaussian Bayes, Discriminant analysis) or non-parametric (Parzen densities, mixture models). These classifiers are called \emph{generative} because they build a probabilistic model of the classification problem that can be used to generate data. 

As a simple example, we consider the Gaussian Naive Bayes (GNB) classifier for a two class classification problem. Training data is  $\{(\x_i,y_i)|i = 1,\ldots,n\}$, where each $y_i$ is either class $1$ or class $2$. For convenience, we define $D_1 = \{i|y_i = 1\}$ and $D_2 =  \{i|y_i = 2\}$, the indices of the training samples belonging to the class 1 and class 2, respectively. Moreover, let $n_1$ and $n_2$ be the number of samples in classes 1 and 2, so that $n = n_1 + n_2$. For GNB, we make the assumptions that 1) data in each class is Gaussian distributed and 2) each feature is independent from each other feature. Denote $\x_i = [x_{i1},x_{i2},\ldots,x_{id}]$, where $d$ is the number of features. The classifier training consists of 
\begin{enumerate}
    \item Computing the mean for each feature $k = 1,\ldots,d$, and each class $j = 1,2$: $m_{jk} = (1/n_j)\sum_{i: y_i = j} x_{ik}$.
    \item Computing the variance for each feature $k = 1,\ldots,d$, and each class $j = 1,2$: $s_{jk} = (1/n_j)\sum_{i: y_i = j} (x_{ij} - m_{jk})^2$.
    \item Computing the estimates $\hat{P}(1) = n_1/n$ and $\hat{P}(2) = n_2/n$.  
\end{enumerate}
After these computations, the class for a test sample $\z = [z_1,z_2,\ldots,z_d]$ can be computed by computing two discriminant values: $$
f_j = \hat{p}(\x|j) \hat{P}(j) = [\prod_{k = 1}^d G(z_k;m_{jk},s_{jk})]n_j/n,$$
and picking the class producing the larger discriminant. The function $G(\cdot;m,s)$ denotes Gaussian probability density with the mean $m$ and variance $s$, i.e., $G(z;m,s) = \frac{1}{{ \sqrt {2 s \pi } }}e^{{{ - \left( {z - m } \right)^2 } \mathord{\left/ {\vphantom {{ - \left( {z - m } \right)^2 } {2s }}} \right. \kern-\nulldelimiterspace} {2s }}}$.

Most modern learning algorithms construct \emph{discriminative} classifiers. These do not aim to construct a generative model for a classification problem, but try to more or less directly find a classification model that minimizes the number of misclassifications in the training data (see \cite{ng2002} for a more elaborate definition of a discriminative classifier). \hlsec{Widely used classifiers such as support vector machines \cite{burges1998tutorial}, Random Forests \cite{breiman2001random}, gradient boosted trees \cite{friedman2001greedy,chen2016xgboost,ke2017lightgbm}, and neural networks \cite{bengio2014deep} belong to this class of classification algorithms.} During the learning process, the classifier is defined by a set of parameter values $\w$, i.e., the classifier is a function $\alpha(\x;\w)$, where the parameter vector $\w$ is to be learned during the training. The training is typically done by optimizing a cost function (or a loss function) $f$ that is related to the desired optimality criterion (for a recent survey of optimization methods in machine learning, see \cite{sun2019survey}). \hl{Note that to make the optimization tractable (for example, to compute the gradients of the loss function), the output of the classifier $\alpha(\x;\w)$ is assumed to be continuous valued instead of the discrete valued. To give a simple example, consider a two-class problem, where the classes are 0 and 1, i.e., $y_i = \{0,1\}$. Then, a widely used loss function is cross-entropy loss, defined as
$$
f(\w) = \sum_{i = 1}^n(y_i \log(p_i) + (1 - y_i)\log(1 - p_i)), 
$$
where $p_i = \alpha(\x_i;\w) \in [0,1]$.  This cross-entropy loss function approximates the number of misclassifications in the training set, which cannot be directly used as a loss function as it not continuous and thus intractable to optimize \cite{NIPS2010_ca8155f4}.} In other words, cross-entropy loss serves as a surrogate loss for the number of misclassifications (or one-zero loss). 

Discriminative classifiers are powerful and often preferred over the generative ones. \hl{They allow for problem-specific loss functions (such as Dice loss or Boundary loss in image segmentation \cite{milletari2016v,kervadec2019boundary})}. \hlsec{Especially, neural networks and gradient boosting machines are very flexible in terms of loss-functions and widely used software libraries allow the user to supply custom loss-functions.} However, for the healthcare applications, where the sizes of training sets are typically small, the formulation of discriminative classifiers poses two problems. First, it may not be clear which optimality criterion the loss function approximates and if the loss function approximates the correct optimality criterion. Second, in the case of separate sampling (which is common in studies in this field), it must be taken into account that the prior probabilities might not be correct \cite{shahrokh2013}. These two problems are less severe with generative classifiers.     

\subsection{Other considerations in classifier design}

\hl{Several choices must be made when designing machine learning based methods for health and wellness applications. In this section, to focus on the essence for this tutorial, we have assumed that we are provided with the feature vectors $\x_i$. In practice, however, it is not the case that we would be directly provided with feature vectors. A traditional view of (general) pattern recognition systems, for example as presented in a classic text of \cite{Duda2000}, divides the system design into five stages (measurement/sensing, pre-processing, feature extraction, classification, and post-processing), out of which classification is just one. Developments in deep learning \cite{goodfellow2016deep,zhang2020dive}, especially convolutional neural networks in imaging applications \cite{ronneberger2015u,tajbakhsh2016convolutional,lundervold2019overview}, have extended learning-based approaches to cover also pre-processing and feature extraction, complementing the traditional engineered (or hand-crafted) features by feature learning \cite{bengio2013representation}. In imaging and genetics, where the number of features often (far) exceeds the number of subjects available, feature selection techniques are essential \cite{guyon2003introduction,saeys2007review,tohka2016comparison}. It is important to bear in mind that all the choices made for the whole system have impact on the final performance and are thus subject to evaluation. However, considerations in this tutorial are largely applicable even when considering the system as whole rather than only the components of the system governed by supervised learning.}

\section{Classifier performance estimates}
The performance of classifiers can be measured in many ways. All these ways are related to the number of times the classifier was 'correct' or 'wrong' when processing and assigning new inputs to classes in real-life usage. However, different usage scenarios introduce different views on what we mean with 'correct' or 'wrong', and (especially in healthcare) not all mistakes are equally important. Thus, we have a range of different performance measures.

Final performance estimates (based on discrete classes) can often not be used directly as optimization criteria for discriminative classifiers, as was explained in Section 2.5. Commonly,  the classifier yields continuous output values which are then thresholded to provide the discrete class-labels that form the basis for performance evaluation. The loss function should approximate the desired performance measure, and often the loss function is just the corresponding performance estimate without thresholding the classifier output.   

\subsection{Simple error rates}

The simplest way of assessing performance is by calculating the number of errors, or, correct classifications, that are made on a given set of data. It is clear that if the number of errors is large the performance is not good. The accuracy can be defined as a simple ratio:

\begin{equation}
\label{eq:accuracy}
accuracy = \frac{\text{number of correct classifications}}{\text{total number of samples to classify}}. 
\end{equation}

A high accuracy (close to 1, or, 100\%) might indicate that we have a useful classifier. However, this is not always certain, especially not if the prevalences of the different groups (e.g., diseased vs. healthy cases) are imbalanced. If, e.g., we have the situation where we have a test test of 1000 cases, with 1 case being a 'disease case' and 999 are 'healthy cases', then we can make a simple classifier that always classifies any case as 'healthy case'. It would have an impressive accuracy of 999/1000 = 99.9\%, but be useless in practice, since its ability to correctly detect actual disease cases (which we call sensitivity) is zero. In healthcare, disease cases are typically much less present than healthy cases, and overall accuracy is thus a poor measure. For this reason, we advance from this simple accuracy measure to a more in-depth look to quantify how well disease (and healthy) cases are classified separately.

\subsection{Confusion matrix}

The confusion matrix is a useful tool for quantifying the performance of a classifier on different classes. Commonly, the rows of the matrix contain the ground truths and the columns the classification results (although it should be noted that this is not set in stone, and allocating the other way around is sometimes used instead). Matrix element $(i,j)$ reflects the number of cases that actually belong to class $i$, and were classified as belonging to class $j$. An example is given in Table \ref{tab:confusion_matrix}.

\begin{table}[]
\renewcommand{\arraystretch}{1.5}
    \centering
    \begin{tabular}{|c|c|c|} \hline
     \diagbox{true class}{classification} & healthy & disease  \\ \hline \hline
      \multirow{2}{*}{healthy}  &  \cellcolor{green} 116  & \cellcolor{red} 5  \\ 
                                &  \cellcolor{green} True Negatives & \cellcolor{red}  False Positives \\ \hline
      \multirow{2}{*}{disease} &   \cellcolor{red} 12 & \cellcolor{green} 23 \\
      &   \cellcolor{red}  False Negatives & \cellcolor{green} True Positives \\ \hline
    \end{tabular}
    \caption{An example of confusion matrix}
    \label{tab:confusion_matrix}
\end{table}

In this case we have a 2-class problem (class 1 = 'healthy', class 2 = 'disease'), and accordingly have a 2x2 matrix. On the rows are the true classes, 'actual healthy' and 'actual disease', as ground truth (e.g., as  observed from a confirmed clinical endpoint/diagnosis), and in the columns are 'classified healthy' and 'classified disease'. We can see that there were 116+5= 121 healthy cases in the set; 116 of them were correctly classified as healthy (we call these True Negatives (TN)), 5 erroneously as disease (‘false alarms’)  (False Positives (FP)). Also, there were 12+23 = 35 disease cases, 23 of them got correctly classified (True Positives (TP)) and 12 wrongly assessed as belonging to the healthy group (False Negatives (FN)). A perfect classifier would have all non-zero values on the diagonal, and zeroes everywhere else. We can see that the classifier overall works quite ok, but especially the detection of disease cases (12 out of 35 classified wrongly) is not so great. Exploring the confusion matrix is thus highly useful to get understanding about what kind of errors are being made on what classes. We can derive several simple quantities from the matrix that succinctly capture its main properties.

\subsection{Classification rate, sensitivity and specificity, FPV, PPV, precision and recall and F1 score}

Performance of classifiers can be quantified by the following measures, see also Table \ref{tab:confusion_matrixy}. All of them have values in the interval $[0,1]$.

{\bf accuracy}: this is the total of correctly classified cases divided by the total number of cases in the test set as in Eq. (\ref{eq:accuracy}). In our example it is 0.891.

{\bf sensitivity}: is the number of disease cases (that are by convention called ‘class positive’) that were correctly classified, divided by the total number of disease cases, i.e., $\frac{TP}{TP + FN}$. It thus quantifies how well the classifier is able to detect disease cases from the disease population, and thus appropriately ‘raises an alarm’. A low sensitivity implies that many disease cases are ‘missed’. In our case the sensitivity is 0.657

{\bf specificity}: is the number of healthy cases (that are by convention called ‘class negative’) that were correctly classified as such, divided by the total number of healthy cases, i.e., $\frac{TN}{TN + FP}$. It thus quantifies how well the classifier is able to detect healthy cases from the healthy population, and thus appropriately ‘stays quiet’. A low specificity implies that many healthy cases are classified as disease case, and many false alarms are generated. In our case the specificity is 0.959.

{\bf positive predictive value (PPV)}: is the number of cases that actually have a disease divided by the number of cases that the classifier classifies as having a disease, i.e., $\frac{TP}{TP + FP}$. It thus is a probability-related measure that indicates how probable it is that a case has a disease when the classifier has a positive/disease class as output. Or, more popularly, how much one should ‘believe’ the classifier when it indicates that the person has a disease. In our case it is 0.821.

{\bf negative predictive value (NPV)}: is the number of cases that actually are healthy divided by the number of cases that the classifier classifies as being healthy, i.e., $\frac{TN}{TN + FN}$. It thus is a probability-related measure that indicates how probable it is that a person is healthy when the classifier has a negative/healthy class as output. Or, more popularly, how much one should ‘believe’ the classifier when it indicates that the person is healthy. In our case it is 0.906.

Sensitivity and specificity are perhaps the most common measures in clinical tests and wider healthcare contexts, when we talk about the performance of (diagnostic) tests or patient monitoring settings. In a wider application area, also the following measures are used.

{\bf Precision}: is the number of cases that actually belong to class X divided by the number of that the classifier classifies as belonging to class X. If we have a two-class classification problem (like in our example), precision is identical to positive prediction value. 

{\bf Recall}: is the number of cases belonging to class X that were correctly classified as class X, divided by the total number of class X cases. In a two-class setting it is identical to sensitivity.
Precision and Recall can be applied to more-than-two class problems, which explains their widespread use in, e.g., reporting of performance of machine learning algorithms. 

The {\bf F1-score} is the harmonic mean of precision and recall: 
\begin{equation}
\text{F1} = 2 \cdot \frac{Precision \cdot Recall}{Precision + Recall}. 
\end{equation}
If either the precision or the recall have small values, the overall F1 score will be small. It thus provides a more ‘sensitive’ measure than accuracy.

\begin{table}[]
\renewcommand{\arraystretch}{1.5}
    \centering
    \begin{tabular}{|c|c|c|c|} \hline
     \diagbox{true class}{classification} & healthy & disease  & \\ \hline \hline
      healthy  &  \cellcolor{green} 116  & \cellcolor{red} 5 & \parbox[t]{3cm}{specificity = \\ $116/(116 + 5)$} \\ \hline
    disease &   \cellcolor{red} 12 & \cellcolor{green} 23 & \parbox[t]{3cm}{sensitivity = \\ $23/(12 + 23)$} \\ \hline
    &  \parbox[t]{3cm}{NPV = \\  $ 116/(116 + 12)$} 
    & \parbox[t]{3cm}{PPV = \\ $23/(5 + 23)$} 
    & \parbox[t]{3cm}{accuracy =\\ $\frac{(116 + 23)}{(116 + 5 + 12 + 23)}$} \\ \hline
     
    \end{tabular}
    \caption{The confusion matrix of Table \ref{tab:confusion_matrix} extended with summary values.}
    \label{tab:confusion_matrixy}
\end{table}

\subsubsection{The effect of class prevalence on PPV and NPV}

A word of caution regarding PPV and NPV: their use is common, partly because of the intuitivity of the measure (“if my classification algorithm says I have a disease, how much should I believe it”). However, it should be kept in mind that PPV and NPV are not only dependent on the performance of the classifier alone, but are also dependent on how many cases of different classes are present in the datasets. This relates to the prevalences of different classes, or prior probabilities, and the earlier example regarding the breast cancer screening (where we ended up with a surprisingly low PPV of 7.8\%) It can be further illustrated with a simple example, see Table \ref{tab:confusion_matrixz1} and \ref{tab:confusion_matrixz2}.

\begin{table}[]
\renewcommand{\arraystretch}{1.5}
    \centering
    \begin{tabular}{|c|c|c|c|} \hline
     \diagbox{true class}{classification} & healthy & disease  & \\ \hline \hline
      healthy  &  \cellcolor{green} 95  & \cellcolor{red} 5 & \parbox[t]{3cm}{specificity = 0.95} \\ \hline
    disease &   \cellcolor{red} 5 & \cellcolor{green} 15 & \parbox[t]{3cm}{sensitivity = 0.75} \\ \hline
    &  \parbox[t]{3cm}{NPV = 0.95} 
    & \parbox[t]{3cm}{PPV = 0.75} 
    & \parbox[t]{3cm}{accuracy = 0.92} \\ \hline
     
    \end{tabular}
    \caption{A dataset with 100 healthy cases and 20 disease cases}
    \label{tab:confusion_matrixz1}
\end{table}

\begin{table}[]
\renewcommand{\arraystretch}{1.5}
    \centering
    \begin{tabular}{|c|c|c|c|} \hline
     \diagbox{true class}{classification} & healthy & disease  & \\ \hline \hline
      healthy  &  \cellcolor{green} 95  & \cellcolor{red} 5 & \parbox[t]{3cm}{specificity = 0.95} \\ \hline
    disease &   \cellcolor{red} 20 & \cellcolor{green} 60 & \parbox[t]{3cm}{sensitivity = 0.75} \\ \hline
    &  \parbox[t]{3cm}{NPV = 0.83} 
    & \parbox[t]{3cm}{PPV = 0.92} 
    & \parbox[t]{3cm}{accuracy = 0.92} \\ \hline
     
    \end{tabular}
    \caption{A dataset with 100 healthy cases and 80 disease cases. Sensitivity and specificity have remained the same, but NPV, and especially PPV have increased considerably.}

    \label{tab:confusion_matrixz2}
\end{table}

The NPV and PPV are influenced by the ratio of disease and healthy cases that happen to be in the test set. If the number of disease cases is high, then also the PPV tends to be high. This is intuitively understandable (if a disease’s prevalence is high, it is easier to believe the classifier when it classifies a case as disease, then it would be when the disease would be rarely occurring). Thus the PPV and NPV are influenced by both the classifier performance {\em and} the number of cases of different classes in the test set. They should thus {\em not be used} to compare classifiers’ performances when those performances have been derived from different datasets. Sensitivity and specificity do not suffer from this problem. 

Different prevalences can (and are likely) to occur when we deal with datasets that have been collected at different centres, in different geopgraphical locations with different processes etc. Another, more technical reason why prevalences may get affected is when training sets are artificially being ‘balanced’ when a certain class is underrepresented. Training a classifier on a class that has only a few instances may be difficult, and one way to deal with that is by repeating/copying the few ‘rare disease’ cases in the set to alleviate training. Thus we are artificially increasing prevalence, and, as a consequence, PPV.

\hl{Many software toolkits for general data analysis and machine learning report \emph{precision} as one of the central measures. As mentioned, for a two-class problem this is equal to PPV. The above discussion should make clear that care should be taken when using this measure to compare performances. Moreover, it may become clear why in healthcare measures like sensitivity and specificity are more often used instead.} 

\subsubsection{Dealing with more than two classes}
\hl{The examples above relate to two-class problems, but in reality we often have cases where we have more than two classes. Examples can be found differential diagnostics of diseases in which the classification task deals with data and subjects that have similar symptoms but different underlying diseases that are hard to discern. Examples can be found, e.g., in the field of dementias, where differentiations need to be made between Alzheimer's dementia, Lewy-body dementia, frontotemporal dementia, and vascular dementia  \cite{tong_five-class_2017, tolonen_data-driven_2018} or patients with tremor symptoms, that may be due to Parkinson's disease, or other tremor-inducing diseases \cite{7367690}. We already saw that precision and recall naturally are defined for any number of classes. Sensitivity and specificity can be generalized to fit to a multi-class problem by grouping classes. In those cases a 2-class setting is generated by classifying one class vs ‘other classes merged’. An example is given in Table \ref{tab:confusion_matrixz_multi}. The numbers are the same as in Table \ref{tab:confusion_matrixz2}, however, the disease-class now has 2 sub-classes. For example, the sensitivity for healthy cases is (vs any disease) is 95/(95+2+3). The sensitivity for disease A is 11/(9+11+19) and so on. In this example it can be seen that the classifier is well able to separate between healthy and diseased cases, but separating between disease A and disease B seems to be problematic. This is a common problem in healthcare applications, where separating between 'healthy' and 'disease' cases is relatively easy (but often clinically less important, since the disease presence is obvious already to the healthcare professional), but classifying between different diseases is a more clinically relevant (and difficult) task, often requiring consulting of (several) experts. The abovementioned examples of dementias and tremor-related diseases are cases in point: separating between a healthy person and a demented person is, from a clinical practice point of view, not a task for which advanced computer-based methods are needed, and neither is separating between healthy subjects and Parkinson's patients. However, help is needed to discern between the subtle differences in the underlying disease groups. 

\begin{table}[]
\renewcommand{\arraystretch}{1.5}
    \centering
    \begin{tabular}{|c|c|c|c|} \hline
     \diagbox{true class}{classification} & healthy & disease A & disease B \\ \hline \hline
      healthy  &  \cellcolor{green} 95  & \cellcolor{red} 2 & \cellcolor{red} 3 \\ \hline
    disease A &   \cellcolor{red} 9 & \cellcolor{green} 11 & \cellcolor{red} 19 \\ \hline
    disease B &   \cellcolor{red} 11 & \cellcolor{red} 15 & \cellcolor{green} 15 \\ \hline
     
    \end{tabular}
    \caption{A dataset with 100 healthy cases and 80 disease cases; disease A has 39 cases, disease B has 41 cases.}

    \label{tab:confusion_matrixz_multi}
\end{table}

}

\subsection{Balanced classification rate}

As we saw, the accuracy-measure does not give a good insight in the overall usefulness of a classifier. Sensitivity, specificity, PPV, and NPV provide better insights. If one wants to summarise the performance in one single number, the balanced accuracy can be used as alternative to the ‘regular’ accuracy. For a 2-class classifier, it can be calculated as the average of sensitivity and specificity. For the general multi-class case it is the average of the proportion of correct classified cases for each class individually. If the classes are ‘balanced’, e.g., there are as many cases in the disease group as there are in the healthy group, then the balanced and regular accuracy are equal. However, in cases where the number of cases for the different classes are not the same (which is very common in healthcare settings), balanced accuracy gives a more appropriate estimation of overall accuracy.

\subsection{ROC curves and AUC}

\begin{figure}
    \centering
    \includegraphics[width=13cm]{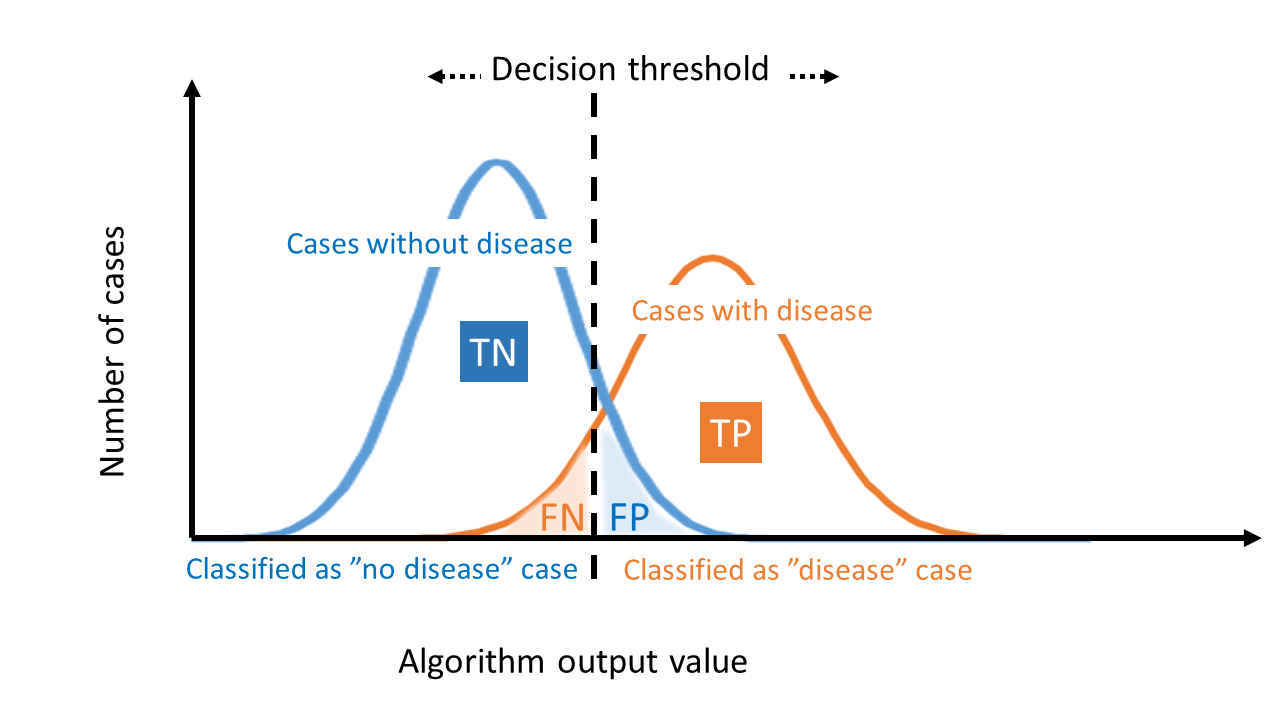}
    \caption{Example distributions of outputs of a classifier algorithm for a 2-class problem. The eventual class assignment is done by using a decision threshold. Moving this value affects how many classified cases are True Negatives (TN), True Positives (TP), False Negatives (FN) and False Positives (FP), and thus, what the values for sensitivity and specificity are.}
    \label{fig:output_distributions}
\end{figure}

In the two-class classification problem there is often a trade-off between having a high sensitivity (detect all persons who have a disease) versus high specificity (avoid false alarms, detect all persons who are healthy). Usually the classification is done by having a classifier output  evaluated and use a threshold on it (or 'decision criterion') to decide whether to assign the input data to class 0 or class 1. The value of the threshold defines then what the values of sensitivity and specificity are. An illustration of the problem is given in Figure \ref{fig:output_distributions}.

\begin{figure}
    \centering
    \includegraphics{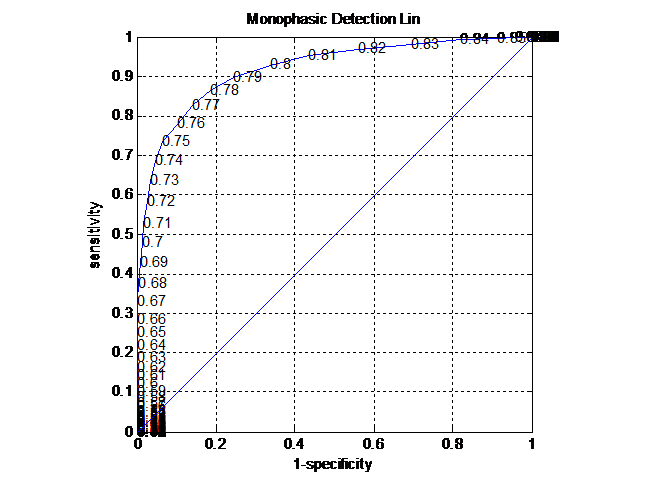}
    \caption{An example of an ROC curve. In this case that of a classifier (Monophasic Linear Detector) that is trained to classify brain activity: monophasic EEG vs ‘normal’ EEG. The numbers along the curve are different thresholds/criterion values to make the final classification. More information can be found in \cite{sarkela2007quantepileptiform} }
    \label{fig:auc}
\end{figure}

If we move the threshold (‘criterion value’ in Fig. \ref{fig:output_distributions}) to low values (and classify all persons who have a classifier output higher than that low value as having a disease), all persons with a disease would be detected (sensitivity = 1). However, also many healthy persons would be classified as having a disease (false alarms, specificity is low). The opposite is also true - a high threshold leads to high specificity but low sensitivity. Selecting the best threshold is thus a trade-off between sensitivity and specificity. The problem can be visualized in the so-called ROC curves (Receiver Operating Characteristic), which have been around since the WWII, but have been introduced into the medical field since the 1970’s. They are a plot of 1-specificity on the x-axis vs sensitivity on the y-axis (in some research disciplines, these axes labels are sometimes labelled as FPR (false positive rate) on the x-axis, and TPR (true positive rate) on the y-axis). An example of an ROC plot is given in Figure \ref{fig:auc}.

It can be seen that any threshold value below 0.72 has a sensitivity smaller than 0.6 and any value above 0.8 has a specificity smaller than 0.6. It can, e.g., be inferred that at  certain places a relatively small increase in threshold (from 0.79 to 0.82) leads to a big effect in specificity (drop from 0.78 to 0.4). This type of explorations helps to make informed decisions on the threshold settings. The point (0,1) would give the ideal classifier, with both sensitivity and specificity having a value of 1. Points on the curve closest to (0,1) would be the ‘best’ classifier, but it has to be kept in mind that usually there is a preference for either the sensitivity or specificity (or both) to be in a certain region (sometimes called 'clinically useful region'), and either of them may be given relatively more importance. Thus, the curve provides a tool to explore the merits of different thresholds.

Another common use for the ROC curve is to calculate its area-under-the-curve (AUC), and use that as performance measure for the classifier. It is a number between 0 and 1, with a value of 1 indicating that the classifier will classify a randomly presented case always correctly. A value of 0.5 indicates the classifier is not better than random guessing (and traditionally the diagonal is plotted in the figure as well, indicating a random classifier for comparison). A useful classifier should have an AUC (significantly) higher than 0.5. How high an AUC should be for it to be ‘good enough’ is application-dependent. For some situations an AUC of 0.7 might already be very good, for others 0.95 might be still rather poor. And, in some cases a lower AUC might be clinically acceptable if, e.g., the sensitivity is high. AUC is convenient because it provides one single number to describe overall performance, and as such, can be used to compare different classifiers against each other. However, it should be kept in mind that it says nothing about the clinical usefulness. A good recent overview of the discussion can be found in \cite{Janssens2020}. In there, it is also described how the ROC curve in fact can be derived from the pdf of the classifier outputs for different classes.

\hl{Although ROC analysis is commonly used for 2-class problems, the idea can also be extended to more-class settings. An approach that extends from the 2-class AUC to a multiclass version, in the form of the volume under the ROC hypersurface (VUS), is presented in \cite{landgrebe_simplified_2006}.
}

\subsubsection{ROC curves and practical utility}
The ROC curve may give us a tool to find optimal threshold values in theory: a point on the curve as close as possible to (0,1) (equal to 1-specificity = 0, sensitivity = 1)). However, there are many trade-offs to make when considering the balance between sensitivity and specificity. For example: the relative costs associated with acting upon a classified disease (when it is a false classification), or the costs associated with not acting in case the disease is actually present. Or, the discomfort to a patient associated to a possible intervention (or discomfort when the disease is not treated) etc. Differences in costs for false positives vs false negatives may give rise to reconsideration of the position of the optimal threshold to choose. Additionally, prevalence of the disease plays a role. A good practical discussion of the ROC in healthcare settings can be found at \footnote{\url{http://www.anaesthetist.com/mnm/stats/roc/Findex.htm}}

There are 3 approaches to find 'optimal' thresholds (see also, \footnote{\url{ http://www.medicalbiostatistics.com/ROCCurve.pdf}}) 
\begin{itemize}
    \item Calculate the point on the ROC curve that has minimum distance to (0,1). This assumes that sensitivity and specificity are of equal importance. It is easy to implement in an algortihm: calculate the distance for each point on the curve, and choose as optimal the threshold the point with the smallest distance.
    \item The second approach uses the logic that the point on the ROC curve that is at the largest vertical distance from the diagonal represents the optimal threshold. Informally, this could be motivated by saying that, since points on the diagonal represent a 'random classifier', points far away from this would represent better classifiers - the further the better. If we consider for a given x-co-ordinate (1-specificity), the y-co-ordinate of the points on the ROC curve the sensitivity, and at the diagonal (1-specificity), then the vertical distance is sensitivity - (1-specificity) = sensitivity + specificity. This is called the Youden index, J. Optimizing this value thus gives a threshold with the best combination of sensitivity and specificity (or a maximum 'balanced accuracy'). Alternatively, it can be looked upon as maximisation of the difference between sensitivity and false positive rate. Again, we assume equal importance for both.
    \item Finally, estimating the optimal threshold based on costs (cost - minimisation) would deliver the value that can be expected to yield the highest benefit in the real-world. This does not assume that sensitivity and specificity are equally important. The issue is that the costs associated to misclassifications are highly diverse and difficult to estimate as they originate from external processes (e.g. treatment processes), personal patient circumstances (co-morbidities, social interactions, occupation), or local  considerations (e.g. costs of tests, reimbursement policies). Costs are either direct or indirect and may be incurred at different time scales. Thus, there is no easy recipe as there was in the first two approaches, and the effort needs to be seen more as part of wider cost-effectiveness and impact-effectiveness studies, which are major disciplines by themselves. Intuitively, it can be understood that, if the cost of missing a disease diagnosis is high, and intervention (even not-needed intervention of a person who is in reality healthy) is safe and cheap, then the best thresholds can be found at the right top-area of the curve: high sensitivity and accepting a high number of false positives. On the other hand, if an intervention carries high-risk and we are not convinced by its effectiveness, the threshold will be in the left bottom corner: we minimise harming non-diseased people, but take missing diseased persons for granted. In many cases in healthcare settings, sensitivity is prioritised over specificity when it comes to detecting critical patient states. However, it is good to keep in mind that false positives are a major burden, e.g., in critical care patient monitoring, leading to 'alarm fatigue' and potentially enormous costs \cite{keller2012clinicalalarmhazards} - minimising false alarms is a main objective in many medical equipment R\&D efforts. Another recent example can be found in the context of antibody testing for Covid-19. A false positive result may wrongly suggest that a person is 'safe' and can interact more freely with others, with potentially disastrous consequences. 
    
\end{itemize}

\subsection{Other performance measures for classifiers}

It is worth mentioning several other performance measures that are commonly used.

\begin{itemize}
    \item 
The Youden index (or Youden's J statistic) is defined as \cite{youden1950index}
\begin{equation}
    J = sensitivity + specificity - 1.
\end{equation}
Expressed in this way, it is obviously equivalent with the balanced classification rate (or, balanced accuracy). However, the Youden index is often used as the maximum potential effectiveness of a biomarker:
\begin{equation}
    J_{max} = \max_c \{ sensitivity(c) + specificity(c) - 1\} , 
\end{equation}
where $c$ is a cut-off point \cite{ruopp2008youden}.

\item The Matthews correlation coefficient (MCC) is the correlation between the observed and predicted classifications \cite{matthews1975comparison} 
\begin{equation}
    MCC = \frac{TP \times TN - FP \times FN}{\sqrt{(TP + FP)(TP + FN)(TN + FP)(TN + FN)}}.
\end{equation}
It takes values between $-1$ and $1$. 

\item The Dice index \cite{dice1945measures} is widely used in the evaluation of image segmentation algorithms as it effectively ignores the correct classification of negative samples (the background region). The Dice index is defined as $ \frac{2TP}{2TP + FP + FN}$. It has close connections to Cohen's Kappa  \cite{zijdenbos1994morphometric} and it is equivalent to Jaccard coefficient (sometimes termed Tanimoto coefficient) \cite{shattuck2001magnetic}.

\end{itemize}

\subsection{Performance measures for regression problems}

In this subsection, we will briefly outline the most important performance measures when facing a regression task, that is, when the variable to be predicted is real valued, instead of categorical. 

The majority of machine learning algorithms for regression problems aim to minimize the mean squared error (MSE):
\begin{equation}
    MSE = \frac{1}{N} \sum_{i = 1}^N (\hat{y}_i - y_i)^2, 
\end{equation}
where $y_i$ is the correct value for target $i$ and $\hat{y}_i$ is our prediction of it. It puts more emphasis on bigger errors more than on smaller ones (which makes sense in many real life applications), it treats positive and negative errors equally (also acceptable in many cases), and the square-function is mathematically convenient for many optimisation algorithms.  A related measure is mean absolute error (MAE):
\begin{equation}
    MAE = \frac{1}{N} \sum_{i = 1}^N |\hat{y}_i - y_i|, 
\end{equation}
which is less 'punishing' to large errors. MSE and MAE report the error in the scale and quantity of the original variables, which makes them sometimes hard to interpret and application dependent. Easier to interpret alternatives are (Pearson) correlation coefficient 
\begin{equation}
    r = \frac{1}{N} \sum_{i = 1}^N \frac{(\hat{y}_i - \hat{m})(y_i - m)}{s\hat{s}}, 
\end{equation}
where $m = (1/N) \sum_{i = 1}^N y_i$ is the mean of the target variables, $\hat{m} = (1/N) \sum_{i = 1}^N \hat{y}_i$ is the mean of the predictions, and $s = \sqrt{(1/N) \sum_{i = 1}^N (y_i - m)^2}$,  and $\hat{s} = \sqrt{(1/N) \sum_{i = 1}^N (\hat{y}_i - \hat{m})^2}$, are the corresponding standard deviations. The correlation coefficient takes values between $-1$ and $1$ and values near or below zero signal useless predictor while high values (near 1) signal very good predictors.

The coefficient of determination (sometimes termed normalized MSE) is defined as $$
Q^2 = 1 - \frac{(1/N)\sum_i (y_i - \hat{y}_i)^2}{(1/N)\sum_{i = 1}^N( y_i - m)^ 2}.
$$ 
Note that in contrast to explanatory modeling, in predictive modeling, the coefficient of determination can take negative values, and it is not equal to correlation squared \cite{moradi2017predicting}. This is why the notation $Q^2$ is recommended instead of $R^2$.  

\subsection{Accuracy of the performance measures}
The above described performance measures (sensitivity, specificity, AUC of ROC etc) all give certain values based on the specific data and algorithm that have been used. How accurate such a value is as estimation of the performance on the general population is an important question. It may be intuitively be expected that a sensitivity of 0.9 as calculated from a dataset with 10000 cases is more accurate than one from 10 cases. Also, how do we compare classifiers, and test e.g., whether algorithm A is 'significantly better' than algorithm B, based on the AUC?

To estimate the standard error in sensitivity and specificity, different approaches exist. They all relate to the calculation of the confidence interval of the binomial proportion (see, e.g., \footnote{\url{https://en.wikipedia.org/wiki/Binomial_proportion_confidence_interval}}). The simplest implementation is the asymptotic approach, which holds if the number of samples is very large. Other, more refined versions use the Wilson score interval or the Clopper-Pearson interval. A convenient calculator for the confidence interval of various measures can be found at the "Diagnostic test evaluation calculator" webpage \footnote{\url{https://www.medcalc.org/calc/diagnostic_test.php}}. 

The confidence interval for the AUC is not trivial to calculate as it requires assumptions about the underlying distributions. Just calculating the mean and standard deviation from a number of pooled AUC observations is not appropriate as the distribution of the ROC is not inherently normal, and the 'samples' (AUC observations) are not independent, since the underlying data remains constant. An authoritive paper from 1982 by Hanley and McNeil \cite{hanley1982meaningROC} gives estimations that are relatively conservative and based on assumptions of exponentialility underlying score distributions. An alternative is to determine instead the maximum of the variance over all possible continuous underlying distributions with the same expected value of the AUC, which gives an unpractically loose estimation. Cortes and Mohri \cite{cortes2004CIforAUC} present an approach which is distribution-independent and thus wider applicable. \hlsec{A commonly used non-parametric method (often referred to as 'DeLong's method') to compare the AUCs of two or more ROCs is presented in \cite{delong_comparing_1988}.}

\subsubsection{Sources of errors}
There can be many reasons why the performance measures are not 'exact'. The abovementioned confidence intervals are based on taking into account natural random variation in the observations, but the variation in the observations can have many underlying reasons, and may not be random at all.

Some reasons why we cannot assess performance measures exactly include:
\begin{itemize}
    \item \emph{Lack of Gold Standards}: in the above discussions we implied that we knew to which class (0 or 1) a person belonged, and what the algorithms 'correct' answer was supposed to be. However, in many cases the situation is not that clear-cut. A 100\% final diagnosis for a form of dementia might be available only once pathology research can be performed once the patient has deceased. Thus, if we classify that person's data while she is alive there may be a change that the 'correct reference' is not actually correct - influencing our performance estimates. For many patient states (e.g., awareness, anesthesia, pain etc.) we have scales that are not absolute but have been accepted as 'good enough' for practical use. All classification estimates on such scales are thus inherently fraught with some uncertainty margin due to the fact that we simply do not know the exact right answer.
    \item \emph{Inter-expert variability}: to develop and train algorithms data needs to be labelled and assigned to classes. This is typically done by experts in the field. In many of the more complex diagnostic there is room for interpretation, expert A might come to a different diagnostic conclusion than expert B (based on earlier experience, processes etc.). In that case the question arises, should we develop a classifier that matches expert A as good as possible, or expert B (or C)? Or take the average of both? For many datasets the reference data labels have been created by having the different experts discuss with each other and come to a comprise labelling that all agree with. Another approach is to quantify the agreement-level between the expert opinions and use that as performance target for the classifier.
    \item \emph{Limited representativeness of the development and test data}: if data has been collected in one hospital only, and is being tested on completely independent data from the same hospital, it may be so that the performance when applied to data from other hospitals is disappointing. Different settings have different practices, different patient populations (with perhaps different disease prevalences), different types of equipment and different staff - this all may lead to drastic changes in the performance measures. Thus, it is essential in many applications to use multi-centre studies involving different hospitals from different countries, to make sure that the performance assessment results are as generally applicable as possible. This, obviously, is an expensive endeavour and a major reason for why uptake of new technologies in clinical practice is slow. 
\end{itemize}

\section{Classifier performance estimation in practice}

In the previous section, we outlined various performance measures. In this section, we describe means to compute the performance measures in practice and discuss potential pitfalls that may arise. We will concentrate on non-parametric estimation principles (validation, cross-validation, bootstrap) that are equally applicable for all the performance measures introduced in the previous section. Hence, for clarity, we will focus on the estimation of the classification error (or equally, $accuracy$ (Eq. (5))), noting that in the most cases, it can be replaced by any performance measure.   

Many fields of biomedicine have published their own guidelines on how to evaluate machine learning algorithms, for example, in radiology \cite{bluemke2019assessing, group2018artificial,kim2019design,park2018methodologic}, and practitioners should be aware of the field-specific guidelines \cite{park2018regulatory}. The TRIPOD (Transparent Reporting of a multivariable prediction model for Individual Prognosis Or Diagnosis) Statement includes a 22-item checklist, which aims to improve the reporting of studies developing, validating, or updating a prediction model \cite{moons2015transparent}. 

At the same time, it needs to be understood that not all the performance estimation tasks are equal: different validation principles may be apt if evaluating the potential of new technology for use in biomedicine or a prototype of a product for clinical use. We recommend reading \cite{van2019predictive}, which summarizes the validation aspects from a clinical viewpoint, making strong points for the public availability of predictive algorithms in health care.   

\subsection{Training error versus test error, holdout method}

Two error types can be distinguished in machine learning: the training error and the test error. The
training error refers to the classification errors for the training samples, i.e., how
large portion of the training samples is misclassified by the designed classifier. The more important error type is the test
error \footnote{Test error is sometimes termed generalization error. However, some authors make a difference between the two terms. See, e.g., \cite{nadeau2003} for an exact definition of the generalization error. }, which describes how large portion of all possible objects is misclassified by the deployed algorithm/model. The theoretical Bayes classifier, introduced in Section 2,  aims to minimize the test error when the class conditional probability densities and priors are known. However, these densities and priors are never known in practice. 

The training error is the frequency of error for the training data. The training
error is an overly optimistic estimate of the test error. For example, the
training error of the nearest neighbour-classifier is automatically zero. Obviously, this is not true
for the test error. The estimation of the test error by the training error is termed
 {\em the resubstitution method}. Using the resubstitution method to estimate classification error is a severe methodological mistake known as "testing on the training data".    

A much better estimate for the test error is obtained by dividing the training
data into two disjoint sets, termed training and test sets. The training set is used  for the training of the classifier
and the test set is solely used to estimate the test error. This method to estimate the true test error is called {\em the holdout method}. If a natural division of the data into test and training sets does not exist, the division needs to be artificially produced. This artificial division needs to be randomized; for example, it is not a good idea to select all the examples of one class as the test set and others as the training set. It is a good practice to make the division stratified, so that there are equal proportions of classes in the training and test sets. Especially in the neural network literature, one encounters divisions of the available data in three sets, termed training, validation, and test. Then, the validation set is used to tune the parameters of the learning algorithm (learning rate, when to stop learning, etc.).   

A hidden difficulty arises when we have several samples obtained from the same subjects collected at different times (see reference \cite{lewis2018t1} for an example how to handle this kind of situation correctly). Then, all the samples obtained from the same subject need to be in either training or test set. It is inappropriate that some samples obtained from subject J (two-years ago) are in the training set while others obtained (a year ago) also from the subject  J are in the test set. In other words, training and test sets must be independent. If the two sets  are not independent, then the estimates of the test error will be positively biased and the magnitude of this bias can be surprisingly large. 


\subsection{Cross-validation}

\begin{figure}[h]
    \centering
    \includegraphics[width=13cm]{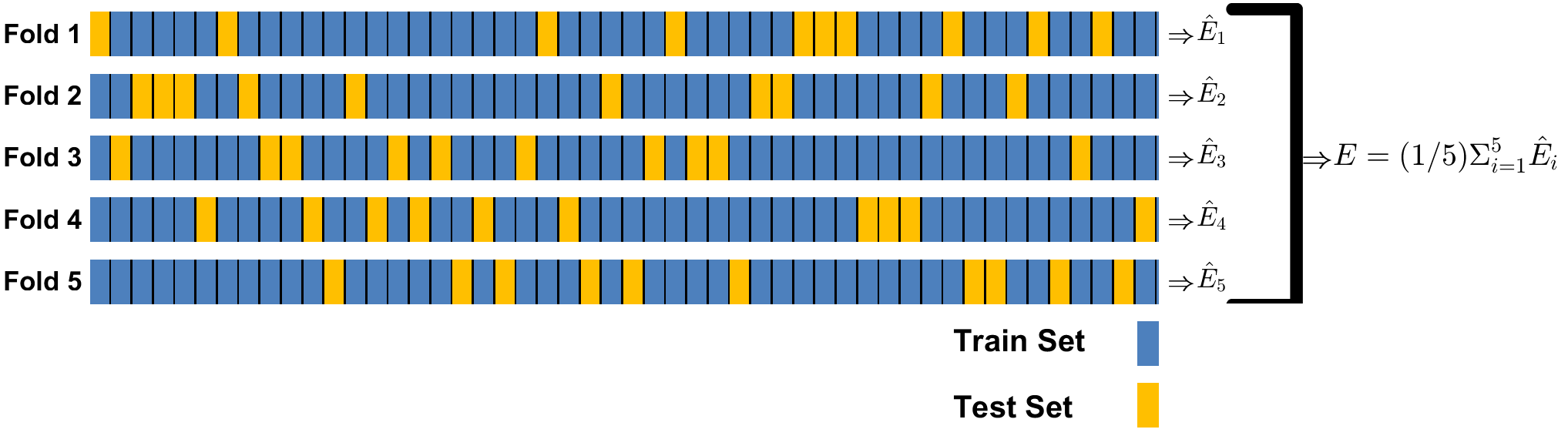}
    \caption{5-fold cross-validation. In each fold, the samples marked by blue color are used as training data, and the samples marked by yellow are the test samples. Each fold produces an error estimate $\hat{E}_i, i = 1, \ldots, 5$, which are then averaged to obtain the total error estimate. 
    }
    \label{fig:cv}
\end{figure}

Cross-validation is a resampling procedure to estimate the test error. It is a generalization of the holdout method. In k-fold cross-validation (CV), the training set $(X,Y)$ is split into $k$ smaller sets $(X_1,Y_1), \ldots, (X_k,Y_k)$ and the following procedure is followed for each of the $k$ folds (see Figure \ref{fig:cv}):

\begin{enumerate}
    \item  A machine learning model is trained using all the except the $i$th fold folds as training data;
    \item the resulting model is tested on $i$th fold $(X_i,Y_i)$.
\end{enumerate}

The performance measure returned by the k-fold CV is then the average of the values computed in the k-folds of the loop. In textbooks that are several decades old, $k$-fold CV has been viewed as computationally expensive, and therefore, the holdout method (see Section 4.1) has been suggested for relatively modest sample sizes. However, computers are now much faster than in the eighties or nineties, and $k$-fold CV is not likely to be computationally prohibitive in current health and wellness applications. Likewise, there exists a wrong perception that holdout would be (theoretically) preferable to CV, but this is not true. Further, the distinction between cross-validation and holdout is not the same as the distinction between internal and external validation. In particular, the holdout method is not the same as having an independent test set. 

The parameter $k$ is usually selected as 5 or 10 according to suggestions given by \cite{kohavi1995study}\footnote{Kohavi \cite{kohavi1995study} recommends to use $k = 10$ as 5-folds led to pessimistic bias in some of the reported experiments}. However, in many cases, there might be a natural division of the data into $k$-folds, for example, the data may have been collected in $k$ different medical centers, and then that natural division should be preferred. A special case, useful when the sample size is small, is leave-one-out CV (LOOCV), where each sample (or subject) forms its own fold, and thus $k$ is equal to the number of data samples. Finally, remarks made about the independence of training and test sets in Section 4.1 hold also for $k$-fold CV, that is, each fold should be independent.

{\em Repeated CV} fixes the inherent randomness of the selection of the folds by re-running a k-fold CV multiple times. There are different opinions if this is useful or not, and in the opinion of the authors, rarely more than ten repeats are necessary. Note that different repeats of CV are not independent, so, for example, the variance of error estimates resulting from different CV runs (note the difference between runs and folds of a single run) is a useless quantity concerning the variance of the generalization error \cite{nadeau2003}. 

In {\em stratified CV}, the folds are stratified so that
they contain approximately the same proportions of labels as in the complete data. For example, if 10 \% of the training data belongs to the class 1 and 90 \% of the training data belong to the class 2, then each fold should also have approximately this 10/90 division. The stratification is typically highly recommended \cite{kohavi1995study}. \hl{In the case of highly imbalanced class proportions, which are discussed in more detail in Section 4.6.2, the stratification is absolutely necessary.}    

\hlsec{An example of a code implementing CV can be found in \url{https://github.com/jussitohka/ML_evaluation_tutorial}. This example uses the voice recordings for the Parkinson's disease detection \cite{naranjo2016addressing}. The data contains three recordings per subject, which must be taken into account when planning the validation.  }  

\subsection{Cross-validation and holdout caveats}
CV and holdout only yield meaningful results if they are used correctly. Common pitfalls include:

\begin{itemize}
    \item Using the test labels (i.e., correct classes of the test set) in feature selection and/or extraction. Selecting features for the classification using the whole data (i.e., not just training set), sometimes termed "peeking effect", leads to optimistically biased classifier performance estimates as demonstrated in, e.g.,  \cite{huttunen2012meg} and \cite{diciotti2013peeking}. However, there exist also more subtle variations of the same issue. For example, preprocessing the data with principal component analysis (PCA) based on all the subjects of the healthy class commits the same mistake as it (inadvertently) uses the labels to select the data for PCA. More generally, if CV is used to simultaneously to optimize the model parameters and estimate the error (within the same CV), the error estimates will be optimistically biased. A procedure called {\em nested CV} is necessary when CV is used and classifier hyperparameters or features need to be selected  \cite{ambroise2002selection,cawley2010over} -- we will return to this issue in Section 4.3.3.  
    \item Failing to recognize that CV-based error estimates have large variance especially when the number of samples is low. In this case,  the classifier may appear very good (or bad) just because of chance.  This issue, discovered over 40 years ago \cite{glick1978additive},  has received a reasonable amount of attention recently \cite{braga2004cross,cawley2010over,hua2009performance,huttunen2015model,rao2008dangers,varoquaux2018cross}, however, its effects are still often underestimated. \hl{Note that the usage of the repeated CV does not cure the problem of increased variance that comes with a small sample size as it only reduces the internal variance of the CV resulting from the random division of samples to different folds.}         
    \item Selecting folds in a way that the training and test sets are not independent, for example, when more than one sample exists from the same subject as we discussed in Section 4.1. 
\end{itemize}

\subsubsection{Fully independent test sets versus cross-validation test sets: is there a difference?}

Many have voiced (e.g., \cite{bluemke2019assessing}) the requirement for independent test sets for the evaluation of machine learning algorithms in health and life science applications.  Especially, if a clinical applicability of a trained machine learning model for a particular task needs to be evaluated, this is absolutely mandatory.  However, we stress that the test set needs 1) to be truly independent (preferably not existing at the training time, see, e.g., a recent competition on Alzheimer's disease prediction for a good example \cite{marinescu2018tadpole,marinescu2020alzheimer}), and 2) needs to model the actual task as well as possible (i.e, collecting the test set at a hospital A when the actual method is to be used at a different hospital B may not be optimal).   

Dividing an already existing dataset into training and test sets as in the holdout method is rarely a good idea if the dataset is not large (in terms of the number of subjects) but it is better to use the cross-validation. What "large" means is dependent on the task at hand \footnote{The question is extremely delicate and parallels the question how much data is needed for machine learning solution for a particular problem. The answer to both questions depends both on the complexity of the task and on the complexity of the selected ML method.}, but to give  a general guideline: 1) datasets of over 10000 subjects can be divided into train and test sets, 2) datasets of 1000 to 10000 subjects, this depends on the case at the hand, 3) datasets under 1000 subjects: cross-validation is typically better. Also, if the data has been collected at several hospitals, it is relevant to perform leave-one-hospital-out cross-validation and report the errors in all the hospitals, instead of selecting some hospitals as training and some as testing sets. 

It is good to keep in mind that the cross-validation approximates the performance of the classifier trained with all the available data. The classifiers derived from using different folds as the training set will differ. Thus, if a specific classifier needs to be evaluated, there is no alternative to the collection of a large test set \cite{van2019predictive}. 

\subsubsection{Cross-validation vs. holdout}

\begin{figure}[h]
    \centering
    \includegraphics{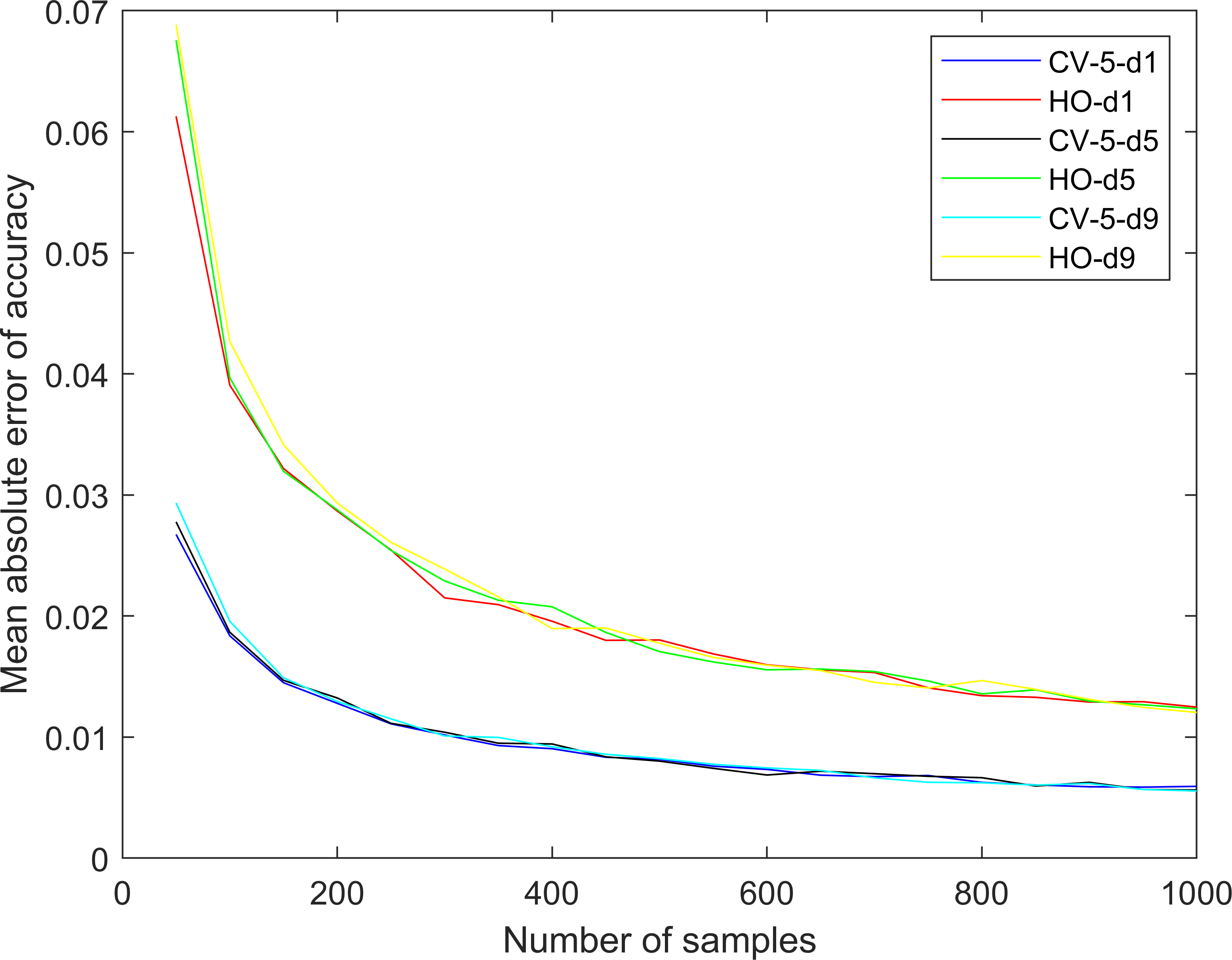}
    \caption{CV versus holdout with simulated data. The plot shows the mean absolute error in the classification accuracy estimation using CV and holdout with respect to a large, external test set consisting of one million samples. The plot shows that the classification accuracy estimate by 5-fold CV is always over two times better than the classification accuracy estimate using holdout with 20 \% of training data in the test set. The number of features was varied ($d=1, 3,5,9$), but the classification task was tuned in a way that the Bayes error was always 5 \%. When the task was tuned this way, the number of features had very little effect on the accuracy of the estimated error.  The classes were assumed to be Gaussian distributed with equal diagonal covariances and the two classes were equally probable.  The classifier was the GNB introduced in Section 2. We trained GNBs with 1000 different training sets of the specific size to produce the figure. The code reproducing this experiment can be retrieved at \protect\url{https://github.com/jussitohka/ML_evaluation_tutorial}}
    \label{fig:cv_vs_holdout}
\end{figure}

Voicing the requirement for independent test sets for the evaluation of machine learning algorithms in health and life science applications has brought with it the confusion that the hold-out would be preferable to the CV-based error estimation. However,  in the absence of a truly separate test set, CV leads always to better estimates of the predictive accuracy than the holdout as is demonstrated by a simulation in Figure \ref{fig:cv_vs_holdout}, where CV-based error estimates have much smaller mean absolute errors than hold-out based ones. This is because, in CV, the results of multiple runs of model-testing (with mutually independent test sets) are averaged together while the holdout method involves a single run (a single test set). The holdout method should be used with caution as the estimate of predictive accuracy tends to be less stable than with CV. In the simulation depicted in Figure \ref{fig:cv_vs_holdout}, both CV and holdout based error estimates are almost unbiased.

\subsubsection{Selecting classifier parameters in holdout or cross-validation}
Modern ML algorithms come often with various hyper-parameters to tune (for example, the parameter $C$ in support vector machines, $mtry$ parameter in Random Forest, when to stop training in neural networks). If using holdout, the recommendation is to divide the data into three non-overlapping sets: 1) a training set to train the classifiers with various hyperparameters, 2) a validation set to decide which of the trained classifiers to use, and 3) a test set to the test the accuracy of the selected classifier. As we have already stated, omitting the test set, and estimating the accuracy based on the best result on the validation set can lead to severely positively biased accuracy estimates.    

Likewise, when CV is used simultaneously for selection of the best set of hyperparameters and for error estimation, a nested CV is required \cite{ambroise2002selection} (sometimes nested CV goes by the name double CV \cite{filzmoser2009repeated}). This again as model selection without nested CV uses the same data to tune model parameters and evaluate model performance leading to upward biased performance estimates demonstrated in several works \cite{ambroise2002selection,huttunen2012meg}. There exist several variants, but the basic idea, as the name indicates, is to perform a CV loop inside a CV loop. First, the whole data set is divided into $k$ folds, and one at the time is used to test the model trained with the remaining $k -1$  folds as in above with the ordinary CV. However, a CV is performed in each of $k - 1$ training sets to select the best hyperparameters, which are then applied to train a model with on the (outer) training sets. A pseudo-code for nested CV is presented in \cite{huttunen2012meg}.        
 
The high variance of the CV (and other non-parametric error estimators) hinders also the selection of hyperparameters for classification algorithms \cite{cawley2010over,huttunen2015model} and, as a result, one should not be overly confident about the selected hyperparameters in small-sample settings.


\subsection{Hypothesis testing}
In order to to compare learning algorithms, experimental results reported
in the machine learning literature often use statistical tests of significance. These tests answer, in a principled manner, the question if one machine learning algorithm is better than another on a particular learning task. However, statistically testing the significance is not completely straight-forward in machine learning scenarios. An early work by Dietterich demonstrated the drawbacks of some commonly used tests and suggested 5x2 CV test and McNemar test  to compare the learning algorithms \cite{dietterich1998approximate}. Nadeau and Bengio proposed corrected resampled t-test, which is the one that we recommend for comparing two learning algorithms \cite{nadeau2003}. This test is advantageous because it takes into account variability due to the choice of training set. A corrected repeated k-fold CV test is a version of this test that is particularly easy to implement \cite{bouckaert2004evaluating}. In this context, we recapitulate that different repeats of CV are not independent and the tests referred here \cite{nadeau2003,bouckaert2004evaluating} take this fact properly into account.

\subsection{Alternatives to cross-validation}
There are several (non-parametric and parametric) alternatives to CV \cite{braga2004bolstered,dalton2010bayesian,efron1997improvements}. We will explain in more detail one of these, bootstrapping \cite{efron1994introduction}, as it is convenient in learning algorithms that utilize  re-sampling (e.g., bagging in Random Forests \cite{breiman2001random}).  In these algorithms, bootstrap (or variations thereof) error estimates are a side-product of the classifier training \cite{breiman1996out}.   

\subsubsection{Bootstrapping}

\begin{figure}
    \centering
    \includegraphics[width=13cm]{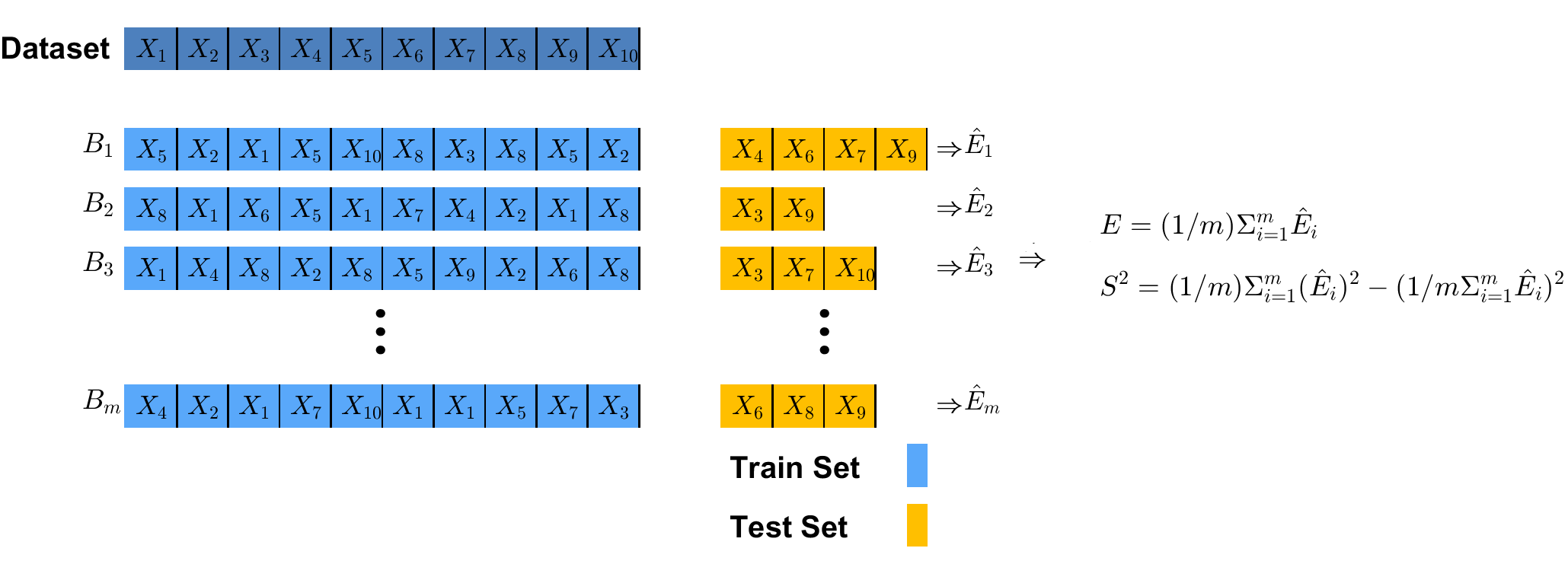}
    \caption{The bootstrap. $m$ samples of the original data are produced by sampling with repetition allowing a training example to enter to the sample more than once. Out-of-bag samples, labeled in yellow in the figure, are then used to estimate the test error. 
    }
    \label{fig:bootstrap}
\end{figure}

In bootstrapping, depicted in Figure \ref{fig:bootstrap}, the central idea is to select random bootstrap samples from the dataset. Typically, these bootstrap samples are of the same size as the original sample, but sampled allowing repetition. (For example, in the first bootstrap sample of Figure \ref{fig:bootstrap} ($B_1$), the sample $X_5$ is selected 5 times.) Then, the classifier is trained with the bootstrap sample and evaluated with the samples that were not selected as part of the bootstrap sample. These samples, labeled with yellow color in Figure \ref{fig:bootstrap}, are called out-of-bag samples. This process is repeated $m$ times. A commonly used variant is .632 bootstrap estimate  \cite{efron1997improvements}, which, however, can be upward biased especially for low accuracies \cite{kohavi1995study}.

\subsection{Practical considerations}

\subsubsection{Dealing with small sample sizes}

A hallmark of healthcare settings is that the sizes of the datasets are rarely large. Collecting data from patients requires addressing technical, organisational as well as regulatory and ethical considerations, and for many diseases their prevalence is low making collection inherently slow. Thus, addressing questions such as "what is the minimum sample size required to design a machine learning algorithm for my problem?" are very common. This is a tricky question as it depends on the application and what is the required performance. Also, for specific applications such as image segmentation, a very small number of images may be sufficient because every pixel is a sample. Instead, image classification may require many more images as now only each subject is a sample. However, the training sample must represent the population of all possible subjects well enough. 

If the available training set size is small in terms of the number of subjects, it is essential to use simple learning algorithms. Note that the term 'simplicity' refers to the number of parameters the algorithm has to learn. For example, a GNB classifier learns $2d + 1$ parameters, where $d$ is the number of features. Nearest-neighbor classifier, albeit simple to implement, leads to much more complicated decision regions and more parameters to be learned  \cite{Hastie2001}. GNB is typically a good choice if the number of samples is small.

Also, as we have emphasized, it is a good idea to take CV-based and, in particular, holdout-based error estimates critically if the number of samples is small. As shown in  Figure \ref{fig:cv_vs_holdout} the mean absolute error of the holdout error-estimate is larger than the Bayes error when the number of samples per class is 25 and, for the majority of iterations, the holdout reports zero errors (while the Bayes error-rate is 5\%). To combat the small sample size problem, there are parametric error estimators that may succeed better than their non-parametric counterparts in small-sample scenarios \cite{dalton2010bayesian}. Note that the effective sample size is the number of training data in the smallest class, i.e., if we have 1000 examples of one class and 3 of another, the trained classifier is not likely to be accurate.

\subsubsection{Dealing with highly imbalanced class divisions}
\label{imbalanced}
\hl{In healthcare applications, it is very common that datasets have (sometimes highly) imbalanced class divisions - diseases, adverse events and emergency situations are by nature less common than the normal/healthy situation. A general overview of the problem and possible approaches can be found in \cite{lopez_insight_2013}. If faced with the situation with highly imbalanced class divisions, few considerations are necessary. First, as outlined in Section 3, it is essential to use an error measure that is appropriate for the problem. Second, if selecting hyperparameters of the algorithm, it is essential to use the same error measure to select the parameter values. Third, the use of stratification in the CV is absolutely necessary. Care should be taken when using oversampling techniques, such as SMOTE \cite{chawla2002smote}, as these do not actually increase the number of training data. Approaches such as RUSBoost \cite{5299216}, combining random undersampling (RUS) with AdaBoost \cite{ratsch_soft_2001} have proven to be successful in healthcare decision support settings where data is imbalanced (e.g., \cite{tong_five-class_2017}). In RUSBoost, training takes place in ensembles of classifiers with iterations using  randomly under-sampled subsets of the data. With each iteration, the weight of the samples is updated: misclassified samples get higher weights and correctly classified samples get less weight. This leads to a situation where misclassified samples will have a better chance of being correctly classified in the next iterations. The final classification is a weighted combination of the entire ensemble's classification outputs. Since samples coming from classes wish a small number of cases are likely to be wrongly classified at the first iteration, they will get increased weights in later iterations and be correctly classified thanks to the boosting process.
It should be mentioned, once more, that for cases where boosting algorithms are used, validation with external datasets or use of the nested CV is essential to obtain reliable performance estimates.} 

\subsubsection{Computing ROCs and AUCs using CV}

There are several choices when combining the AUCs from the different test partitions. Two of the simplest are \cite{bradley1997use}:
\begin{itemize}
    \item Pooling. The frequencies of
true positives and false positives are averaged. In
this way one average, or group ROC curve is
produced from the pooled estimates of each point
on the curve. 
    \item Averaging. AUC is calculated for each test partition and these AUCs are then averaged.   
\end{itemize}
More techniques for combining ROCs from several test partitions are introduced in \cite{fawcett2006introduction}.

\subsubsection{Data augmentation}

\hl{Data augmentation is a technique to increase the diversity of the training set by applying random (but realistic) transformations. For example, in medical image analysis, these transformations might be rotations and contrast transformations \cite{abdollahi2020data}. There exist also general techniques to automatically augment the training data \cite{cubuk2019autoaugment,lim2019fast}. For the performance assessment perspective, it is important to note that data augmentation should only be applied to the training set (and in the case of CV, only to the training folds). That is,  data augmentation is a step in the training pipeline, which comes after splitting your data into train and test sets. Otherwise, test and training data are not independent leading to positive bias in the performance assessment.}   

\hlsec{
\subsubsection{Illustrative examples}

To conclude this section, we summarize two representative example studies to show how confusion matrices, different performances measures, and their uncertainty measures can be reported in practice. A specific example of a multi-class classification problem is presented by Tong and co-workers in  \cite{tong_five-class_2017}. It dealt with the classification of subjects into five different classes of dementia (Alzheimer's dementia, frontotemporal lobe degeneration, dementia with Lewy bodies, vascular dementia, and persons with subjective memory complaints who served as healthy controls) using multi-modal data. The study had access to a large number of cases ($N=500$), but there was a considerable class imbalance. To address the class imbalance, Tong et al. used the RUSBoost approach as discussed in Sec. \ref{imbalanced}. They evaluated the classifiers in terms of accuracy (75.2\%) and, in this case, more relevant, balanced accuracy (69.3\%). Based on 10-fold cross-validation results, they demonstrated that RUSBoost was more effective than, e.g., Support Vector Machines, multi-class cost learning k-nearest neighbors, or Random Forest-based approaches. 

Another example of comparison of the performances of different ML-based classification methods was presented in \cite{hernesniemi_extensive_2019}. The classification task was related to the prediction of mortality in acute coronary syndrome (ACS). The explored classification methods were logistic regression and extreme gradient boosting. The performance assessment focused on the comparison of AUCs of ROC-curves between the different classifiers and also against the traditionally used GRACE score \cite{dascenzo_timi_2012}. The method used to compare the AUCs was DeLong's non-parametric test \cite{delong_comparing_1988}. The results demonstrated that ML methods with extensive data as input have the strong potential to outperform currently standard approaches, such as the GRACE score. Thanks to the good data-availability ($N=9066$ in total), the authors used independent training and test sets in this study. They developed the classifiers using data from a training set of patients treated in 2007 -– 2014 and 2017 (81\%, $N_{train}=7344$) and tested in a separate test set of patients treated in 2015 -– 2016  (19\%, $N_{test}=1722$).


}

\section{Conclusions}
The increasing efforts in research and development using data-driven approaches have both positive and negative effects. On the positive side, new applications and solutions are delivered to problems that were ten years ago still considered prohibitively difficult to solve. Think, e.g., of image recognition, speech recognition, natural language processing in general, and the highly successful advances in medical technology, especially in the image analysis field and assisted diagnostics. On the negative side, there is a hype situation in which there are unrealistic expectations with regard to techniques such as AI and machine learning. High expectations are set both by end-users or customers as well as in the scientific community itself. This leads to a situation where a thorough objective assessment of the performance is a task that is under pressure. As we have seen, proper performance assessment is not trivial (requires an understanding of the problem and data), and often costly (data needs to be reserved) and costs time. It is often less exciting than the development/training work itself and has the unfortunate property that often the performance estimations after objective validation are less than the 'highly promising' results that were obtained in the early development phase. Thus, meaning that enthusiasm from colleagues, potential customers, investors, and end-users might decrease as a consequence of 'proper' testing. All in all, there is pressure to deliver fast and publish (positive!) results as soon as possible.    
The tendency to publish only results that 'improve upon the state-of-the-art (SotA)' is an increasing problem. This problem leads to situations where algorithms and parameter sets are tuned and re-tuned almost indefinitely towards an as good as possible performance; 'SotA-hacking' \hl{(see, e.g., discussions in \cite{DBLP:journals/corr/abs-1904-07633, Rogers_2020_reviewing-models})}. This becomes highly problematic when the overall dataset is fixed, as is the case in publicly available databases or datasets available for pattern recognition competitions. No matter whether an appropriate CV scheme is used and an 'independent' test set is provided, the fact remains that an enormous amount of effort is dedicated to finding the optimal solution to one specific data set. There is insufficient information about how the algorithm behaves on other data in real-life. This is sometimes referred to as 'meta-training', as researchers are getting trained themselves and their environment to optimise their work for a specific data-set. 

The overall problem is wide and ultimately originates from inappropriate experimental design and hypothesis testing procedures, including so-called \emph{Hypothesizing After the Results are Known} (aka HARKing) practices. The interested reader can find more information in \cite{DBLP:journals/corr/abs-1904-07633}

This paper aimed to give practical information about how to assess the performance of machine learning approaches, giving special attention to healthcare applications. Whereas the emphasis here is on machine learning applications, the reader is encouraged to take a broader look at established statistical approaches and findings regarding clinical prediction models in general, e.g., as thoroughly treated in the work by Steyerberg \cite{Steyerberg_ClinicalPredictionModels}. 

Finally, it is worthwhile to re-emphasise that good performance is only one of the items in the list of requirements towards successful uptake of machine learning algorithms in practice: usability, seamless integration into existing processes and infrastructures, and explainability are all components that are of similar importance. They have their own metrics.

\section*{Declarations of interest}
Declarations of interest: none

\section*{Acknowledgments}

J. Tohka's work has been supported in part by grants 316258 from Academy of Finland and S21770 from European Social Fund. The funders had no role in study design, data collection and analysis, decision to publish, or preparation of the manuscript.

We thank Vandad Imani, University of Eastern Finland for drafting the figures \ref{fig:cv} and \ref{fig:bootstrap}. 

\section*{References}





\bibliographystyle{model1-num-names}
\bibliography{sample.bib}









\end{document}